\pdfoutput=1

\documentclass[11pt]{article}

\usepackage[]{acl}
\usepackage{times}
\usepackage{latexsym}

\usepackage[T1]{fontenc}

\usepackage[utf8]{inputenc}

\usepackage{microtype}

\usepackage{graphicx}
\usepackage{caption}
\usepackage{subcaption}
\usepackage{microtype}
\usepackage{tabularx}
\usepackage{multirow}
\usepackage{booktabs}
\usepackage{graphicx}
\usepackage{ragged2e}
\usepackage{footnote}
\usepackage{bbm}
\usepackage{bm}
\usepackage{latexsym}
\usepackage{amsmath}
\usepackage{algorithmic}
\usepackage[ruled,vlined]{algorithm2e}

\newcommand\edu{$^\heartsuit$}
\newcommand\cs{$^\spadesuit$}
\newcommand\psy{$^\clubsuit$}

\title{\textsc{SETSum}: Summarization and Visualization \\ of Student Evaluations of Teaching}

\author{\cs Yinuo Hu\thanks{\,\, Equal contribution.} $\;\;\;$ \cs Shiyue Zhang\footnotemark[1] $\;\;\;$ \psy\edu Viji Sathy $\;\;\;$ \psy\edu A. T.  Panter $\;\;\;$ \cs Mohit Bansal \\
  \cs Department of Computer Science, UNC Chapel Hill  \\
  \psy Department of Psychology and Neuroscience, UNC Chapel Hill \\
  \edu Office of Undergraduate Education, UNC Chapel Hill \\
  {\tt \{huyinuo, shiyue, mbansal\}@cs.unc.edu;} \\
  {\tt \{viji.sathy, panter\}@unc.edu}
}

\begin{document}
\maketitle
\begin{abstract}

Student Evaluations of Teaching (SETs) are widely used in colleges and universities. 
Typically SET results are summarized for instructors in a static PDF report. The report often includes summary statistics for quantitative ratings and an unsorted list of open-ended student comments. The lack of organization and summarization of the raw comments hinders those interpreting the reports from fully utilizing informative feedback, making accurate inferences, and designing appropriate instructional improvements. 
In this work, we introduce a novel system, \textsc{SETSum}, that leverages sentiment analysis, aspect extraction, summarization, and visualization techniques 
to provide organized illustrations of SET findings to instructors and other reviewers. Ten university professors from diverse departments serve as evaluators of the system and all agree that \textsc{SETSum} helps them interpret SET results more efficiently; and 6 out of 10 instructors prefer our system over the standard static PDF report (while  the remaining 4 would like to have both). This demonstrates that our work holds the potential to reform the SET reporting conventions in the future.
\end{abstract}

\section{Introduction}
\label{sec:intro}

Colleges and universities rely on student evaluations of teaching (SETs) to assess students' perceptions about their courses \cite{chen2003student, zabaleta2007use}.
These evaluations about the course consist of both quantitative ratings using Likert-type scales and open-ended comments that describe student experiences. 
In many universities, SETs are a standard component of evaluations of teaching and have multiple functions. First, they help individual faculty members examine their own teaching performance in a diagnostic way so they can work to improve their approach in subsequent offerings of the course. 
Second, SETs allow institution leaders to review 
and describe the educational quality of course offerings and the performance of instructors. 
Third, though controversial, SET summaries often are  used as part of an instructor's larger portfolio to demonstrate their teaching history during high-stakes settings.
Finally, in some colleges and universities, SET summaries are released to students to help guide them with course selections. Given this wide range of uses for the SET summaries, it is important that thoughtful, accurate, and well-designed representations are provided to draw accurate inferences about teaching
quality, course design, and student learning (see ethics Sec.~\ref{sec:ethical} for more details).

Usually, at the end of each semester, SET results are summarized into a PDF report for instructors or other reviewers. As shown in a sample standard SET report in Fig.~\ref{fig:usualSET} of Appendix, 
quantitative ratings are summarized using basic statistics, such as mean and median, while students' comments from open-ended questions are simply listed as raw text – without adequate organization and analyses. When a college course is particularly large (e.g., with more than 100 students), the final SET report can be longer than 10 pages, which is time-consuming to read and analyze \cite{alhija2009student}.  In addition, instructors’ or other reviewers’ own cognitive biases may lead to inaccurate inferences and analyses, e.g., people tend to pay more attention to negative than positive comments \cite{kanouse1987negativity}.

Therefore, the goal of our work is to provide a new dynamic presentation of SET results to facilitate more efficient and less biased interpretations compared to the standard PDF report. After obtaining institutional SET data of four semesters from the University of North Carolina (UNC) at Chapel Hill,
for demonstration we select two quantitative and two open-ended questions from the total number of questions (Sec.~\ref{sec:data}). We develop a system, \textsc{SETSum}, to summarize and visualize the results of these four questions. For quantitative ratings (Sec.~\ref{sec:rating}), 
we visualize two statistics: \emph{response rate} and \emph{sentiment distribution}.
For open-ended comments (Sec.~\ref{sec:comment}), we develop a sentiment analysis model to predict whether each comment sentence is positive or negative. 
We use an aspect extraction approach to help instructors quickly know the popularly discussed topics by students, e.g., assignment, and the corresponding topic sentiments. 
Finally, we propose an unsupervised extractive summarization method that extracts top sentences with high centrality, low redundancy, and balanced sentiments as a summary of each aspect. 

Automatic evaluations (Sec.~\ref{sec:auto_eval}) demonstrate that our sentiment prediction and aspect extraction modules achieve good accuracy, and our summarization method produces more diverse and less biased summaries than simply picking top central sentences. More critically, the effectiveness of \textsc{SETSum} should be judged by its main users -- instructors. Thus, we begin by conducting human evaluations (Sec.~\ref{sec:human_eval}) with 10 
professors from 8 different academic departments at UNC. Note that \textsc{SETSum} is continuously under development, and our human evaluations were conducted on our very first version: \textsc{SETSum v1.0}. After evaluating the two SET presentation approaches, instructors are asked to complete a survey comparing the usefulness of \textsc{SETSum} to the standard SET report. According to their responses, most of the new features introduced on \textsc{SETSum} are perceived as \emph{useful} to \emph{very useful} by most instructors (on average, 8.8 out of 10), compared to the standard report. All 10 instructors agree that \textsc{SETSum} helps them interpret their ratings and comments more efficiently; while 4 out of 10 think the new system also supports less biased interpretations. Finally, 6 of 10 favor \textsc{SETSum} more than the standard approach; the remaining 4 think both reports could be helpful. Overall, for our first evaluation, instructors hold a \emph{positive} attitude towards \textsc{SETSum} and offer valuable and constructive suggestions to us. 

Lastly, in Sec.~\ref{sec:ethical}, we discuss if machine-involved representations of SETs may introduce new errors or bias and if so, what improvement needs to be made before the ``demonstration'' can transition to an ``application''.  
Our system aims to provide accurate, efficient, and visualized SET results to instructors or other reviewers. It does not directly make any value judgments or evaluations about the instructor's skills, the course design, or the amount of student learning during the term. 

To the best of our knowledge, despite its widespread use, we are among the few researchers to develop a pilot system that presents student-reported evaluations of teaching 
by using natural language processing (NLP) techniques. In addition, we are the first to apply the system for a SET instrument and evaluate it using actual SET data from a large public university. Though more development work is in progress, our results demonstrate that our approach is promising to reform the SET report conventions in the future. Our \href{https://setwebsite.netlify.app/}{\textsc{SETSum v1.1}} website requires credentials to login, please contact us for an access to the website. We provide a \href{https://youtu.be/-Z2BBS7dvw0}{YouTube video} to walk you through \textsc{SETSum v1.1}. Our code is hosted at \href{https://github.com/evahuyn/SETSum}{SETSum Github Repo}.\footnote{SETSum v1.1: \url{https://setwebsite.netlify.app/}, YouTube: \url{https://youtu.be/-Z2BBS7dvw0}, Github: \url{https://github.com/evahuyn/SETSum}} 

\begin{figure*}
    \centering
    \includegraphics[width=0.99\textwidth]{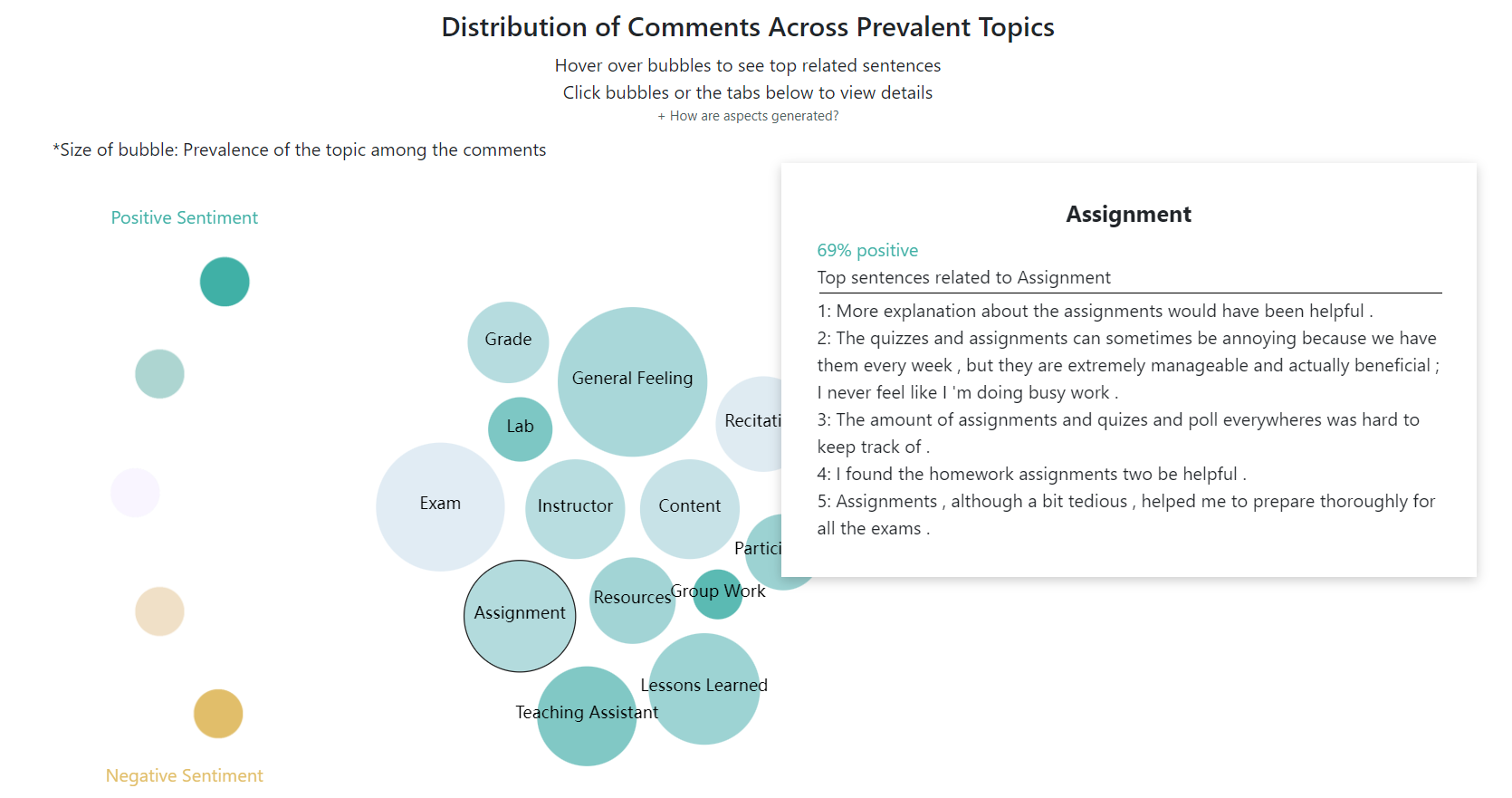}
    \vspace{-5pt}    \caption{The distribution of comments across prevalent topics (aspects). On the left, it shows the aspect bubble chart, and on the right, it shows the summary of the ``assignment'' aspect.} 
    \label{fig:bubble}
    \vspace{-10pt}
\end{figure*}

\section{Background \& Related Work}
\label{sec:related}
As mentioned, SETs are widely used in higher education \cite{chen2003student, zabaleta2007use}. 
SET studies have shown that they can capture students' opinions about instruction \cite{balam2010student}, enhance course design, can be used as a tool for assessing teaching performance \cite{penny2004effectiveness, chen2003student}, and reflect institutional accountability about teaching \cite{spooren2013validity}. Many instructors view SETs as valuable feedback to improve their teaching quality \cite{griffin2001instructor, kulik2001student}. Many studies focus on instrument design (i.e., which questions to ask),  reliability and validity of SET results (i.e., are the scores consistent across contexts; are scores related to other key constructs), and potential confounding variables that affect SETs (e.g., do scores differ by discipline, instructor race/ethnicity and gender, student grade) \cite{simpson2000student, spooren2013validity}.

Typical SET instruments include quantitative Likert-scale ratings. They are supplemented by open-ended comments \cite{stupans2016student, marshall2021contribution}.
Therefore,  compared to quantitative ratings, open-ended comments are often under-analyzed or ignored completely due to labor required to provide an adequate summary \cite{alhija2009student, hujala2020improving}, raising the need for contemporary methods in automated text analysis. Recent works start to analyze student comments via text mining and machine learning methods such as sentiment analysis \cite{wen2014sentiment, azab2016analysing, cunningham2018visually, baddam2019student, sengkey2019implementing, hew2020predicts}, and identify \emph{topics}, \emph{themes}, or \emph{suggestions} from student comments \cite{ramesh2014understanding, stupans2016student, gottipati2018text, unankard2019topic, hynninen2019distinguishing}. The common goal of these works is to answer some research questions (e.g., what are sentiment differences across courses and students).
In contrast, we provide a demonstration tool of SET results to help instructors gain insights on their own and to allow others have access to organized summaries. 

The most relevant works to ours are SUFAT \cite{pyasi2018sufat} and Palaute \cite{gronberg2021palaute} -- two analytic tools for student comments. They both support sentiment analysis and LDA topic models \cite{blei2003latent}, while we use more advanced RoBERTa-based sentiment analysis and weakly-supervised aspect extraction models. SUFAT requires users to install the tool and load SET files locally, while our online website directly reads the data from the SET instrument. 
More importantly, none of them conducts human evaluations, which makes it unclear if their tools are useful from the actual users' 
perspectives. Therefore, we develop the first demonstration system that uses an actual university SET instrument and is evaluated by university instructors who are interpreting their own evaluations. 

\section{SET Data}
\label{sec:data}

We use Student Evaluations of Teaching (SETs) data of over four academic terms (Fall 2017, Spring 2018, Fall 2018, Spring 2019) collected at UNC Chapel Hill. We utilize ``semester + course number'' as the specific identity of each course. We assume each course has just one instructor.\footnote{This is not always true, and we will deal with co-teaching situations in the next version of our system.} 
In total, there are about 5.6K courses and 298K SETs. Each SET is an evaluation questionnaire assigned to a student for a specific course they enrolled in, including both quantitative and open-ended questions. 

UNC's SET instrument includes a series of evaluation questions assessing different aspects of the course and instructor. For demonstration, we select four representative questions -- two quantitative and two open-ended items. For quantitative items, we choose \textit{Overall, this course was excellent (Course Rate)} and \textit{Overall, this instructor was an effective teacher (Instructor Rate)}, showing students' overall ratings on the course and instructor performance. Both items are based on a 5-point Likert scale (1 = Strongly Disagree to 5 = Strongly Agree).
For open-ended items, we choose \textit{Comments on overall assessment of this course (Course Comments)} and \textit{Comments on overall assessment of this instructor (Instructor Comments)}. Our system can be easily extended to the full set of SET questions. 

Because completing the SET form is not mandatory, the average response rates of the two quantitative items we choose are 46\% and 43\% respectively, and even lower response rates are observed for the two open-ended items: 17\% and 16\%, respectively. 

\section{System Description}
After logging in, instructors can select to display their SET results of which semester and which course. 
The dashboard shows two main sections: \emph{Rating Analysis} and \emph{Comment Analysis}. See screenshots of our demo in Fig.~\ref{fig:SETSumWebv1} (\textsc{SETSum v1.0}) and Fig.~\ref{fig:SETSumWebv2} (\textsc{SETSum v1.1}) in Appendix. 

\subsection{Rating Analysis}
\label{sec:rating}
For each of the two quantitative questions, we show the following statistics.

\paragraph{Response Rate.}
Since students do not always respond to every SET question,
knowing how many students responded is critical for interpreting the generalizability and representativeness of the results. The standard report (Fig.~\ref{fig:usualSET}) provides the number of responses for each question. To make this information stand out, we use a circular packing chart to describe the proportion of students who answered the question in comparison to the total enrollment of the course (Fig.~\ref{fig:response_rate}).  

\paragraph{Sentiment Distribution.}
The standard SET report summarizes quantitative ratings by mean, median, standard deviation, and percentages of the 5 rating options. Instead, in \textsc{SETSum v1.0}, we simplify ratings to be either \textit{positive} (4 and 5) or \textit{negative} (3 or lower). 
We show the positive vs. negative ratings via a pie chart (Fig.~\ref{fig:sentiment} in Appendix). 
However, after conducting human evaluations on \textsc{SETSum v1.0}, we received feedback from instructors preferring the original 5-point scale distribution. Thus, in \textsc{SETSum v1.1}, we include a detailed breakdown of all scores. 

\subsection{Comment Analysis}
\label{sec:comment}
For open-ended questions, besides the option to view all raw comments as the standard report (by clicking the ``View Raw Comments'' button in \textsc{SETSum v1.0} or the ``Table View'' button in \textsc{SETSum v1.1}), we provide the following new features.

\subsubsection{Basic Statistics} 
We present the \textbf{Response Rate} of open-ended questions also by a circular packing chart.
Student comments are raw texts without sentiment labels. Therefore, we develop a sentiment analysis model (Sec.~\ref{sec:sentiment}) and get the sentiment of each comment sentence. Then, we display the \textbf{Sentiment Distribution} (positive vs. negative ratio) via a pie chart for instructors to acquire an overview of students' sentiments expressed in comments. 

\subsubsection{Sentiment Analysis} 
\label{sec:sentiment}
As mentioned in Sec.~\ref{sec:related}, many existing works have conducted sentiment analysis on SET data \cite{wen2014sentiment, azab2016analysing, baddam2019student}. In UNC's SET instrument, no sentiment labels are explicitly related to student comments.
To train a sentiment analysis model, we pair 
\textit{Course Comments} with the 
\textit{Course Rate} since they are both overall assessment of the course. Similarly, we pair 
\textit{Instructor Comments} with \textit{Instructor Rate}.

We want to get sentence-level sentiments to compute the overall sentiment of each aspect (Sec.~\ref{sec:aspect-ext}) and conduct summarization (Sec.~\ref{sec:summarization}). However, ratings are comment-level sentences. Thus, we first train a comment-level sentiment analysis model, and then we use it to predict the sentiments of each comment sentence.

\subsubsection{Aspect Extraction} 
\label{sec:aspect-ext} 
Students usually comment on some common \emph{aspects} of the course, e.g., grade, assignment. Previous works resort to LDA \cite{blei2003latent} to learn \emph{topics} from student comments  \cite{ramesh2014understanding, pyasi2018sufat, gronberg2021palaute}. We argue that each topic learned from LDA is a set of words that is hard to be assigned a post hoc name, and topics sometimes lack distinctions \cite{ramesh2014understanding}. Therefore, we apply a weakly-supervised aspect extraction model, MATE \cite{angelidis-lapata-2018-summarizing}, that can extract aspects from comments using a set of pre-defined aspects.

\paragraph{MATE.} \textit{Multi Seed Aspect Extractor} (MATE) \cite{angelidis-lapata-2018-summarizing} is derived from \emph{Aspect-Based Autoencoder} (ABAE) \cite{he2017unsupervised}. ABAE learns a sentence-level aspect predictor without supervision by reconstructing the sentence embedding as a linear combination of aspect embeddings. Assume $\mathbf{v}_s$ is the sentence embedding and $\mathbf{A}$ is a matrix of aspect embeddings, ABAE first predicts aspects:
$\mathbf{p}_s^{aspect} = softmax(\mathbf{W}\mathbf{v}_s+\mathbf{b})$, and then reconstructs the sentence vector: $\mathbf{r}_s = \mathbf{A}^\top\mathbf{p}_s^{aspect}$. The objective is a max-margin loss using random sentences $n_i$ as negative examples:
$$\mathcal{L} = \sum_s\sum_i max(0, 1 - \mathbf{r}_s\mathbf{v}_s + \mathbf{r}_s\mathbf{v}_{n_i})$$
\noindent Similar to LDA, ABAE has to interpret the learned aspects post hoc. To address this, MATE pre-defines a set of aspects by humans, and each aspect is given a set of \emph{seed words}. Concatenating seed word embeddings together forms an aspect seed matrix $\mathbf{A}_i$, and the final aspect embedding matrix $\mathbf{A}=[\mathbf{A}_1^\top \mathbf{z}_1, ..., \mathbf{A}_K^\top \mathbf{z}_K]$, where $\mathbf{z}_i$ is a weight vector of seed words. 

\paragraph{Aspect Annotation.} To pre-label aspects of student comments and get seed words for each aspect, we randomly sample 100 comments for each of the two open-ended questions from the entire corpus and split them into sentences. Two human annotators (two authors) work together, attribute one or more aspects to each sentence, and label the corresponding aspect sentiments (positive or negative). Table~\ref{tab:annotation} in Appendix shows two examples. In the end, we obtain 15 and 11 aspects of comments on course and instructor, respectively, and their terminology is defined in Table~\ref{tab:terminology_cou}, \ref{tab:terminology_ins} in Appendix. With the annotations, we calculate \textit{clarity} scores \cite{cronen2002predicting} of each word w.r.t. each aspect (see details in Appendix~\ref{sec:app-clarity}). 
The higher the clarity score, the more likely the word will appear in sentences of a specific aspect. We manually select 5 top-scored words for each aspect while removing noise (stopwords, names). Their scores are re-normalized to add up to 1. Table~\ref{tab:aspects} shows the 5 seed words (plus weights) for each aspect.

\paragraph{Visualization.} After training the MATE model, we predict the aspects of each comment sentence. We select all aspects that have $p_s^{aspect} > 0.4$. The threshold (0.4) is tuned on the subset with aspect annotations.  Then, for each open-ended question of each course, we visualize its aspect distribution via a bubble chart (Fig.~\ref{fig:bubble}). Bubble size represents the number of sentences of this aspect. While bubble color denotes the aspect sentiment -- the average of sentence-level sentiments, we chose accessible color palette (the more blue the more positive, the more yellow the more negative).

\subsubsection{Extractive Summarization}
\label{sec:summarization}

After clustering comments by aspects, we want to provide a summary of each aspect. We first obtain the ``centrality'' of each sentence and then propose a method to extract summaries with high centrality, low redundancy, and balanced sentiments.

\paragraph{LexRank.} For all the comment sentences under a certain aspect, we use LexRank \cite{erkan2004lexrank} to get the graph-based ``centrality'' of each sentence, where we use the cosine similarity of sentence embeddings from Sentence-BERT \cite{reimers2019sentence}.
Intuitively, if a sentence is similar to many other sentences, it will be close to the ``center'' of the graph and thus it is prominent.

\begin{algorithm}[t]
    \small
    \SetAlgoLined
    \KwIn{$S^a, K$ }
    \KwOut{$S'$} 
    $S'  \gets \emptyset$,~~$S'_a \gets S_a$,~~$k  \gets 1$\;
    \While{$k \leq K$}{
        $s \gets \arg \max_{s \in S'_a} J(s, S', S_a)$\;
        
        $S' \gets S' \cup \{s\}$\;
        $S'_a \gets S'_a - \{s\}$\;
        $k \gets k+1$
    }
    \caption{Summarization}
    \label{alg:summ}
\end{algorithm}

\paragraph{Sentence Extraction Algorithm.} Naively, we could extract the top central sentences as the summary. However, such summary sometimes includes redundant information and tends to only select positive sentences as they are more common. Inspired by \citet{hsu2021decision}, we propose a greedy sentence extraction algorithm that optimizes three objectives on sentence selection: (1) maximizes centrality;
(2) maximizes the difference between the sentence and other sentences extracted from previous steps;
(3) minimizes the difference between the summary sentiment and the overall sentiment of the aspect.
Algorithm~\ref{alg:summ} demonstrates our unsupervised extractive summarization algorithm, in which $S_a$ represents all sentences under an aspect $a$, $K$ is the number of sentences we want to extract (K=5), and $S'$ is the target summary. Our learning objective (we want to maximize it) at each extraction step is written as:
\begin{align*}
    J(s, S', S_a) = \text{centrality}_s & - \text{cosine\_sim}(s, S') \\ -  \text{senti\_diff}&(S'\cup\{s\}, S_a)
\end{align*}

\noindent Essentially, we want to extract a summary with high centrality, low redundancy, and a balanced sentiment. $\text{centrality}_s$ is the centrality of sentence $s$. Following \citet{hsu2021decision}, we define $\text{cosine\_sim}(s, S')$ as follows:
$$\text{cosine\_sim}(s, S') = \max\limits_{s' \in S'} \text{cosine}(v_s, v_{s'})$$
where $v_s$ and $v_{s'}$ are sentence embeddings from Sentence-BERT \cite{reimers2019sentence}. And we define $\text{senti\_diff}(S'\cup\{s\}, S_a)$, as the following:
$$\text{senti\_diff} = |\frac{\sum_{s' \in S'\cup\{s\}}p(s')}{|S'\cup\{s\}|} - \frac{\sum_{s' \in S_a}p(s')}{|S_a|}|$$
where $p$ is the probability of positive sentiment predicted by our sentiment analysis model.

\paragraph{Visualization.} 
Hovering any bubble in the aspect bubble chart will display its summary on the right (Fig.~\ref{fig:bubble}). 
Clicking on the aspect tab will display listed summary sentences within their the original comments to provide contextual information. A table of all sentences is on the bottom. 
\section{Evaluation \& Results}

\subsection{Automatic Evaluation}  
\label{sec:auto_eval}

\paragraph{Sentiment Analysis.} We train two comment-level sentiment analysis models for \textit{Course Comments} and \textit{Instructor Comments} respectively. We split our data into training (90\%) and development (10\%) sets, and about 6.3K and 5.8K examples are in Course and Instructor development sets respectively. We first 
report comment-level sentiment prediction performance on the dev sets. Second, we use the comment-level sentiment analysis models to predict sentence-level sentiments during inference. To evaluate this, we use our aspect annotation data (Table~\ref{tab:annotation} in Appendix),
and we only use sentences with just one sentiment (i.e., all aspects are positive or negative), resulting in 202 and 230 testing examples for Course and Instructor. We report micro F1 (=accuracy) and macro F1. Table~\ref{tab:sentiment_res} shows the results. It can be seen that our models achieve reasonably good sentiment prediction performance, though they perform worse on predicting sentence-level sentiments than the comment level.

\paragraph{Aspect Extraction.} 
Similarly, we also train two aspect extraction models for \textit{Course Comments} and \textit{Instructor Comments} separately. We evaluate their performance by comparing to human annotated aspects using F1 score. In total, we have 213 and 234 testing examples for course and instructor models, and the average number of aspects is 1.38 and 1.31, respectively. We achieve F1 score of 48.6 for the course model and 48.9 for the instructor model, which are similar to the results of the MATE paper \cite{angelidis-lapata-2018-summarizing}. We also explore another approach by treating aspect extraction as a multi-label aspect classification task. We use half of the annotated data to finetune a RoBERTa-base~\cite{liu2019roberta} model and test on the other half annotated aspects. Our experiment shows improved F1 scores of 62.6 for the course model and 64.9 for the instructor model. We plan to combine RoBERTa and MATE to deploy a weakly-supervised RoBERTa-based MATE in our next version of website. 

\paragraph{Summarization.} Due to the lack of gold summaries, we use three metrics (\emph{Centrality}, \emph{Redundancy}, and \emph{Sentiment Difference}) to evaluate our summarization approach and compare it to the baseline of extracting the top 5 central sentences. Please refer to Appendix~\ref{app:summ-evel} for detailed definitions of these three metircs.
We randomly sampled 100 courses as the testing set to report the performance. Table~\ref{tab:summarization_res} shows the results. As expected, our method leads to lower redundancy and sentiment difference than the baseline, though it scarifies some centrality. 

\begin{table}
\centering
\small
\resizebox{0.43\textwidth}{!}{%
\begin{tabular}{lcc}
\toprule
Sentiment Analysis & Course & Instructor\\
\midrule
Comment-level micro F1 & 0.87 & 0.94 \\
Comment-level macro F1 & 0.83 & 0.86 \\
\midrule
Sentence-level micro F1 & 0.83 & 0.90 \\
Sentence-level macro F1 & 0.84 & 0.85 \\
\bottomrule
\end{tabular}
}
\vspace{-7pt}
\caption{Sentiment analysis results.}
\label{tab:sentiment_res}
\vspace{-7pt}
\end{table}

\begin{table}
\centering
\small
\resizebox{0.43\textwidth}{!}{%
\begin{tabular}{lcccc}
\toprule
Summarization & \multicolumn{2}{c}{Course} & \multicolumn{2}{c}{Instructor} \\
\cmidrule(lr){2-3} \cmidrule(lr){4-5} 
& Base. & Ours & Base. & Ours \\
\midrule
Centrality$\uparrow$ & \bf 1.13 & 1.09  & \bf 1.14 & 1.10\\
Redundancy$\downarrow$ & 0.05 & \bf 0.03 & 0.05 & \bf 0.02 \\
Sentiment Diff$\downarrow$ & 0.41 & \bf 0.34 & 0.43 & \bf 0.36\\
\bottomrule
\end{tabular}
}
\vspace{-7pt}
\caption{Summarization results.}
\label{tab:summarization_res}
\vspace{-13pt}
\end{table}

\subsection{Human Evaluation}  
\label{sec:human_eval}
It is critical to evaluate how our demonstration system is perceived by its primary users: instructors. 

\subsubsection{Evaluation Setup} 

\paragraph{Design a Survey.} We design an evaluation survey using Qualtrics. Our complete survey can be found at \href{https://github.com/evahuyn/SETSum/blob/main/survey/SETSum_survey_and_results.pdf}{SETSum Github Repo}. In the survey, we first introduce the background and purpose. We define the standard PDF report  \textit{Usual Approach} and our \textsc{SETSum v1.0} as \textit{Comparison Approach}, and then we ask instructors to compare the two approaches. The main survey body contains 5 parts of questions:

(1) 
\emph{Rate the Usual Approach}: Without comparing to \textsc{SETSum}, we ask how they rate the usefulness of standard SET reports in a 5-point scale: not at all, slightly, moderately, very, or extremely useful; 

(2) 
\emph{Rate \textsc{SETSum} (Rating Analysis)}: Compared to the usual approach, instructors rate our new features of summarizing ratings in a slightly different 5-point scale: not at all useful, not useful, equally useful, useful, or very useful; 

(3) 
\emph{Rate \textsc{SETSum} (Comments Analysis)}: Compared to the usual approach, we ask how useful each of our new features of summarizing comments is (using the same response anchors as  (2)).

(4) \emph{Rate the overall experience with \textsc{SETSum}}: We ask if our website helps them interpret SET results more efficiently and/or with less bias (definitely not, probably not, might or might not, probably yes, definitely yes) as well as if they prefer the standard SET report or our website or both. 

(5) \emph{Comments}: Instructors may leave additional comments on the website under development.

\paragraph{Invite Instructors.} We invited 15 professors at UNC, who taught large introductory courses within the studied period (4 semesters). We estimated the survey to take 20-30 minutes, and each participant was offered a \$25 gift card to a campus coffee shop.  
In the end, 10 instructors from 8 different departments completed the survey successfully.

\subsubsection{Results Analysis}

Fig.~\ref{fig:HumanEval} shows the survey results, and the Qualtrics report can be found at \href{https://github.com/evahuyn/SETSum/blob/main/survey/SETSum_survey_and_results.pdf}{SETSum Github Repo}. Here, we summarize some main takeaways.

\paragraph{Instructors have positive opinions about the standard SET report.} 8 out of 10 (and 6 out of 10) instructors think the PDF report is \emph{moderately to extremely useful} in summarizing students' ratings (and comments), respectively. This demonstrates the well-perceived usefulness of existing SET reports by instructors, though they are less satisfied with the comment summarization.

\paragraph{New features introduced on \textsc{SETSum} are perceived to be useful or very useful.} 
On average, for rating analysis, 7 out of 10 instructors think each of the 2 new features (response rate and sentiment distribution) is \emph{useful} or \emph{very useful}, and for  comments analysis, 8.8 out of 10 instructors on avg. think each of the 5 new features (response rate, sentiment distribution, topic bubbles, summary sentences, showing original comments for each summary sentence) is \emph{useful} or \emph{very useful}, while fewer instructors (5.5 out of 10 on avg.) think the scatter plot\footnote{We had a scatter plot showing all comment sentences in \textsc{SETSum v1.0}, which was removed from \textsc{SETSum v1.1}.} and the table showing all comment sentences are \emph{useful} or \emph{very useful}.
Overall, most instructors perceive our \textsc{SETSum} as being useful.

\paragraph{\textsc{SETSum} helps all instructors interpret SET results more efficiently, and it helps some instructors interpret SET results with less bias.} All instructors agree that \textsc{SETSum} helps them interpret SETs \emph{more efficiently} (i.e., probably to definitely yes). 4 out of 10 instructors think it helps them understand SETs \emph{with less bias}. 

\paragraph{Instructors prefer \textsc{SETSum} than the standard report or would like to have both.} Lastly, 6 out of 10 instructors prefer \textsc{SETsum} compared to the usual approach, while 4 instructors would like to have both approaches. 

\paragraph{Constructive suggestions.} We identify the following suggestions from instructors' comments for improving our future version: (1) The accuracy of the sentiment analysis and aspect extraction models can still be improved. 
(2) Many instructors prefer the complete display of ratings in the 5-point scale, rather than presenting only a positive v.s. negative ratio.
(3) Instructors without a computer science background had difficulty understanding concepts like ``centrality''. 
So far, we addressed (2) and (3) in \textsc{SETSum v1.1} by providing the 5-point scale rating distribution and adding detailed explanations for each Machine Learning related modules. 
\vspace{4pt}

\noindent Overall, instructors show a very positive attitude towards our \textsc{SETSum} demonstration system and provided important suggestions and direction for our future work.

\section{Conclusion}
In this work, we propose \textsc{SETSum}, a system to summarize and visualize results from student evaluations of teaching. We integrate NLP, statistical, visualization, and web service techniques. We are among the few researchers to build a tool for instructor use and are the first to evaluate the tool by university professors. Our results demonstrate that our system is promising at improving the SET report paradigm and helps instructors gain insights from their SETs more efficiently. In the future, we will keep improving the sentiment analysis and aspect extraction models to provide more accurate summarization of SET results. The instructor evaluation offered key recommendations for the next iterations of the system. We will incorporate more functions to our system, including allowing instructors to compare different courses and track their own teaching history of their courses as well as developing a separate administrator dashboard to identify themes across academic courses, departments, and programs. 

\section{Ethical Considerations}
\label{sec:ethical}

As mentioned earlier, SETs have multiple functions such as (1) faculty members examining their teaching performance in a diagnostic way, (2) allowing institution leaders to review and describe the quality of course offerings, (3) part of an instructor's larger portfolio to demonstrate their teaching history during high-stakes settings, and (4) summaries being released to students to guide them with course selections. Given this wide range of uses for the SET summaries, our work's purpose is to take initial steps towards developing thoughtful, accurate, and well-designed representations that can be provided to draw accurate inferences about teaching quality, course design, and student learning. However, it is also critical to examine all aspects of SETs through an ethical lens. Errors in NLP-based analysis could lead to misinterpretation and inaccurate judgments in high-stakes settings. Therefore in the following, we discuss how each module of our SETSum website affects the interpretation of SET results.

For quantitative items, we provide visualizations of two statistics that are directly computed from SET data. Therefore, no errors or biases should be introduced compared to the standard PDF report. In fact, some instructors who participated in our evaluations say that the \emph{response rate} feature for each individual question helps them understand the results with less bias.

For open-ended items, to obtain their sentiment distributions, we develop sentiment analysis models to obtain sentence-level sentiments. Though we obtain good sentiment prediction performance (Table~\ref{tab:sentiment_res}), errors are inevitable. We use these features to demonstrate the relative number of positive to negative comments (ratio). In general, unless very few students evaluate a course (low response rate for comments), the system can still convey the information fairly well. Another important feature that we develop as part of this system is to group comment sentences by aspects. Although we achieve similar aspect prediction F1 scores consistent with past research, we find that the results are not precise enough yet for widespread use. Our human evaluators notice that some sentences from the open-ended comments are inaccurately clustered. Therefore, in future iterations of this system, we believe it is very important to develop a more accurate aspect extraction model. The final important feature is the unsupervised extraction summarization module. We choose an extraction method because it does not suffer from faithfulness (not staying true to the source) issues as abstractive methods \cite{cao2018faithful}. Meanwhile, our algorithm extracts summaries with more balanced sentiments (Table~\ref{tab:summarization_res}). Nonetheless, we hope to find a summarization approach that aligns more closely with the sentiments underlying the students' comments.

Though instructors express positive attitudes towards our system and 4 instructors think it help them understand SETs with less bias, we believe that additional thorough evaluations need to be conducted in the future. Outside of SETs, our work recognizes the different ways, reporters, and methods that could be used to assess teaching effectiveness, including but not limited to peer reports, analysis of classroom sound, student learning, and an instructor's own teaching portfolio.

Finally, at this time our system is designed to be used and reviewed by instructors or other reviewers, and it \emph{does not} directly make any broad judgments or decisions (e.g., whether the instructor is qualified for promotion). The primary end-users of the system should be instructors who wish to analyze their SET findings more thoroughly and acquire the main takeaways more efficiently. Other reviewers and administrators can use the system to view the SET findings in a broader scope, such as reading the report summary per department or per division.
Overall, the goal of \textsc{SETSum} is to help instructors and other reviewers to understand more of students' needs and make improvements to future course design. 

\section*{Acknowledgments}
We thank the reviewers for their helpful comments. We thank the UNC instructors who participated in our human evaluations. We would also like to thank Heather Thompson from the Office of Undergraduate Curricula for providing the data and Rob Ricks from the Office of Institutional Research and Assessment for additional data preparation activities. This work was supported by NSF-CAREER Award 1846185, NSF-AI Engage Institute DRL-2112635,  a Bloomberg Data Science Ph.D. Fellowship, and the Howard Hughes Medical Institute Inclusive Excellence 3 Grant. 

\bibliography{anthology,custom}
\bibliographystyle{acl_natbib}

\appendix

\section*{Appendix}
\label{sec:appendix}

\begin{figure}
    \centering
    \includegraphics[width=0.40\textwidth]{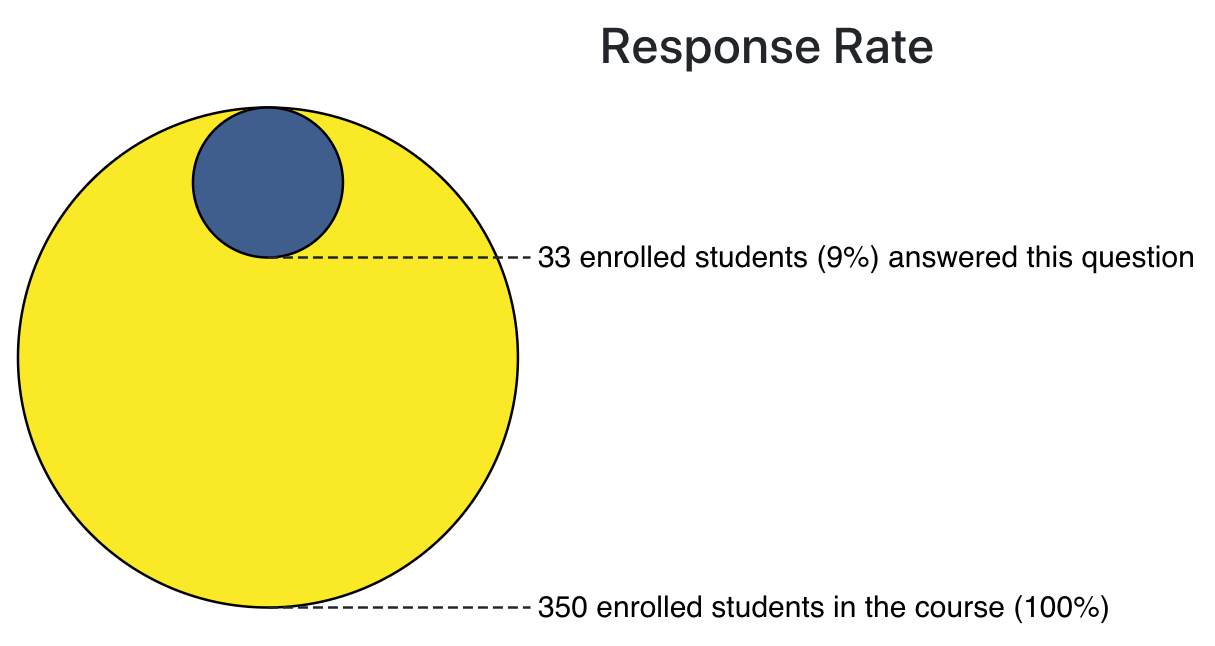}
    \caption{A circular packing chart describes the response rate.}
    \vspace{-7pt}
    \label{fig:response_rate}
    \vspace{-7pt}
\end{figure}

\begin{figure}
    \centering
    \includegraphics[width=0.40\textwidth]{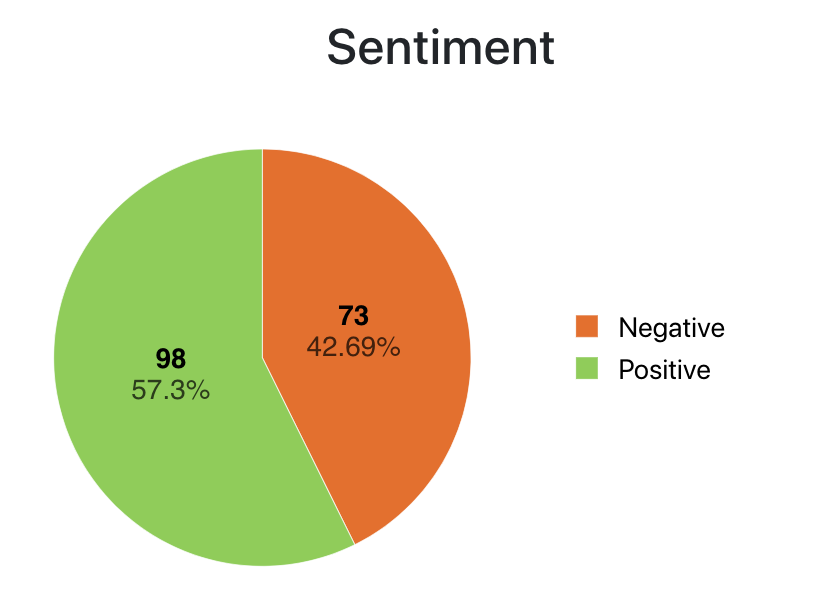}
    \caption{A pie chart describes the sentiment distribution.}
    \vspace{-7pt}
    \label{fig:sentiment}
    \vspace{-7pt}
\end{figure}


\section{Clarity Score}
\label{sec:app-clarity}
To identify seed words for each aspect. We compute the clarity score \cite{cronen2002predicting, angelidis-lapata-2018-summarizing} of each word with respect to each aspect. The score measures how likely it is to observe word $w$ in comments of aspect $a$: $score_a(w) = t_a(w)log\frac{t_a(w)}{t(w)}$, where $t_a(w)$ is the tf-idf score of $w$ in comments of aspect $a$ and $t(w)$ is that in all comments. 

\section{Implementation Details}
\label{sec:app-imp}
\paragraph{Sentiment Analysis.} We finetuned a RoBERT-large model \cite{liu2019roberta} using HuggingFace's Transformers \cite{wolf-etal-2020-transformers} for 5 epochs and chose the best performed checkpoint on the development set. We used the AdamW optimizer \cite{loshchilov2018decoupled} with learning rate 1e-5 and batch size 16. 

\paragraph{Aspect Extraction.} We used NLTK to conduct sentence and word segmentation. We initialized the MATE model using GloVe embeddings \cite{pennington2014glove}. During the training, the word embeddings, seed word matrices, and seed weight vectors were fixed and we trained the model for 10 epochs using Adam optimizer \cite{kingma2015adam} with learning rate $10^{-1}$ and batch size 50. We also experimented the multi-label classification approach by finetuning a RoBERTa-base model \cite{liu2019roberta} for 10 epochs using AdamW optimizer \cite{loshchilov2018decoupled} with learning rate 2e-5 and batch size 16. 

\paragraph{Website.} We developed the website using the React framework for the front-end interface. For the back-end, we set up a database with Firebase and created a RESTful API with Firebase Cloud Functions. Our website is deployed to Netlify.com for an online demonstration.

\section{Summarization Evaluation Metrics} 
\label{app:summ-evel}
We define the \emph{Centrality} metric as the average centrality of summary sentences. The higher the metric is, the better. Assume the summary to evaluate is $S'$.
$$\text{Centrality}(S') = \frac{\sum_{s \in S'} \text{centrality}_s}{|S'|} $$
We compute the information \emph{Redundancy} within a summary $S'$ by taking the average of cosine similarity among sentences. We use sentence embeddings from Sentence-BERT \cite{reimers2019sentence} to compute cosine similarities. The lower the metric is, the better.
$$\text{Redun}(S') = \frac{\sum_{s \in S'} \max\limits_{s' \in S' - \{s\}} \text{cosine}(v_s, v_{s'})}{|S'|} $$
We first compute the average sentiments for the summary $S'$ and all sentences under the aspect $S_a$, respectively. Then, we take their absolute difference as the final score of \emph{Sentiment Difference}. The lower the metric is, the better.
$$\text{Senti\_diff} = |\frac{\sum_{s \in S'}p(s)}{|S'|} - \frac{\sum_{s \in S_a}p(s)}{|S_a|}|$$
where $p$ is the probability of positive sentiment predicted by our sentiment analysis model.

\begin{table*}
\centering
\small
\resizebox{0.99\textwidth}{!}{%
\begin{tabular}{p{0.25\textwidth}|p{0.5\textwidth}|p{0.2\textwidth}}
\toprule
\textbf{SET Question} & \textbf{Comment Sentence} & \textbf{(Aspect, Sentiment)}\\
\midrule
\textit{Comments on overall assessment of this course} & Because, even though the lecture was fine the exams were brutal or was just wrong because of the answer key being wrong. & (content, positive); (exam, negative) \\
\midrule
\textit{Comments on overall assessment of this instructor}& The instructor was clear at explaining information and fairly evaluating all assignments. & (delivery, positive); (grade, positive) \\
\bottomrule
\end{tabular}
}
\vspace{-7pt}
\caption{Two examples of Aspect Annotation.}
\label{tab:annotation}
\vspace{-12pt}
\end{table*}

\begin{figure*}
     \centering
     \begin{subfigure}[b]{\textwidth}
         \centering
         \includegraphics[width=0.92\textwidth]{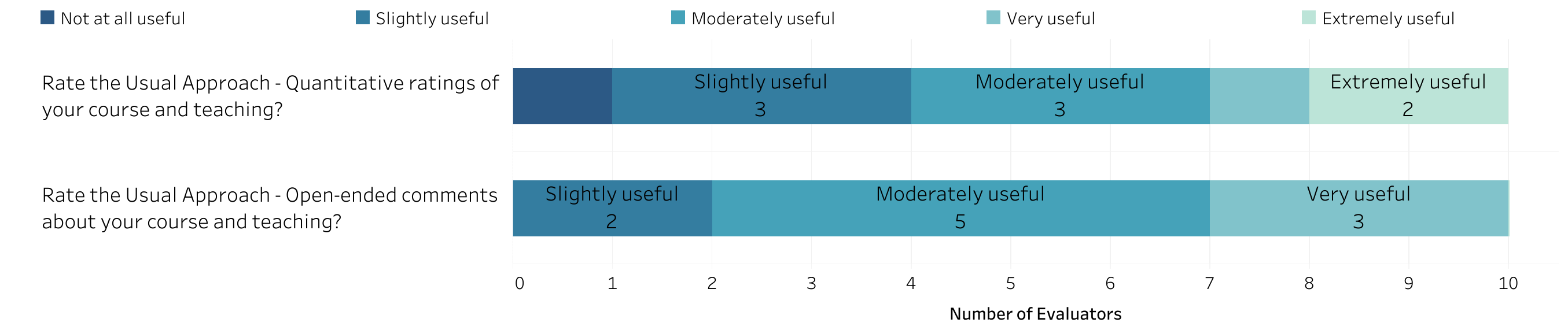}
         \subcaption{\textit{How useful is the \textbf{Usual Approach} in summarizing students' quantitative ratings and open-ended comments of your course and teaching?}}
         \label{fig:Q1-breakdown}
     \end{subfigure}
     \hfill
     \begin{subfigure}[b]{\textwidth}
         \centering
         \includegraphics[width=0.92\textwidth]{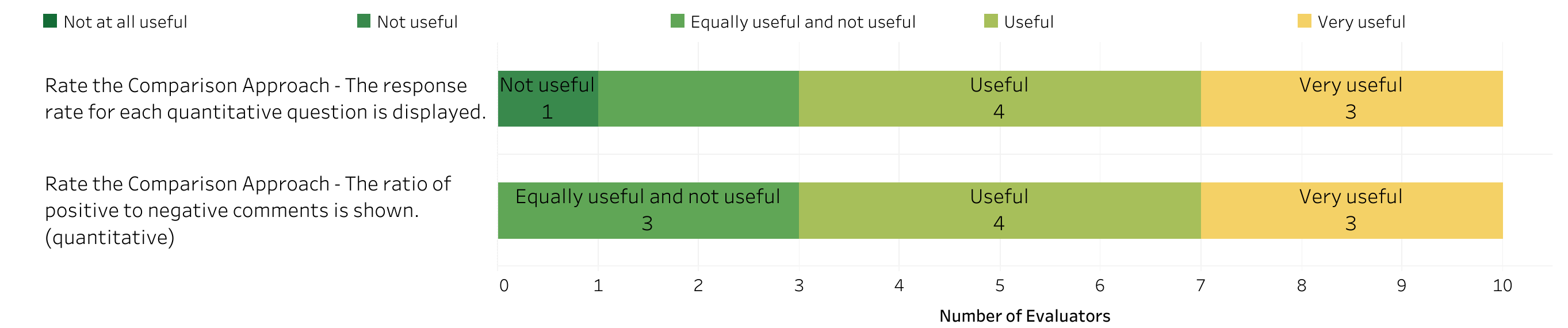}
         \subcaption{\textit{How useful is the \textbf{Comparison Approach} in summarizing students' quantitative comments about your course and teaching?}}
         \label{fig:Q2-breakdown}
     \end{subfigure}
     \hfill
     \begin{subfigure}[b]{\textwidth}
         \centering
         \includegraphics[width=0.92\textwidth]{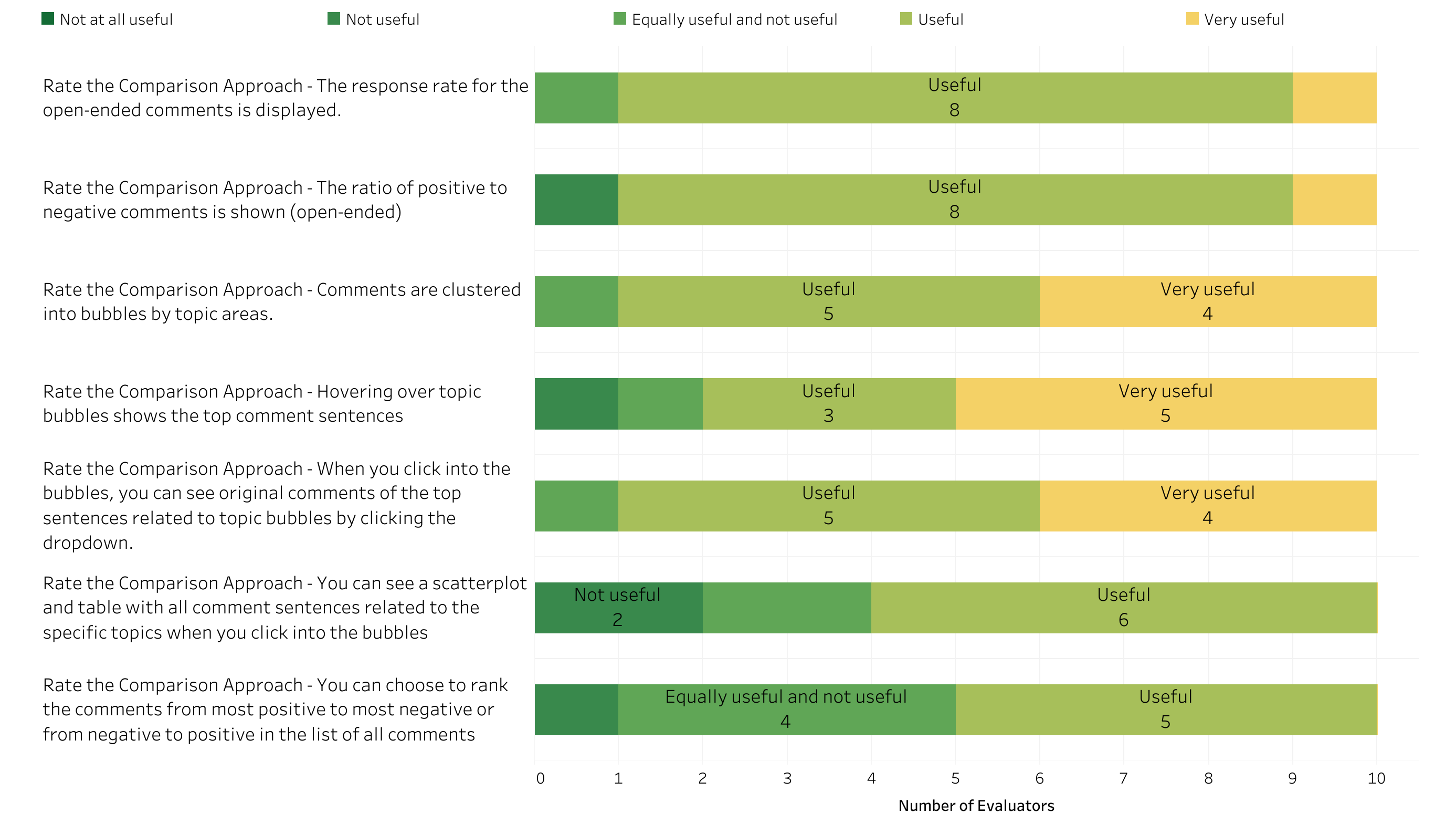}
         \subcaption{\textit{How useful is the \textbf{Comparison Approach} in summarizing students' open-ended comments about your course and teaching?}}
         \label{fig:Q3-breakdown}
     \end{subfigure}
     \hfill
     \begin{subfigure}[b]{\textwidth}
         \centering
         \includegraphics[width=0.92\textwidth]{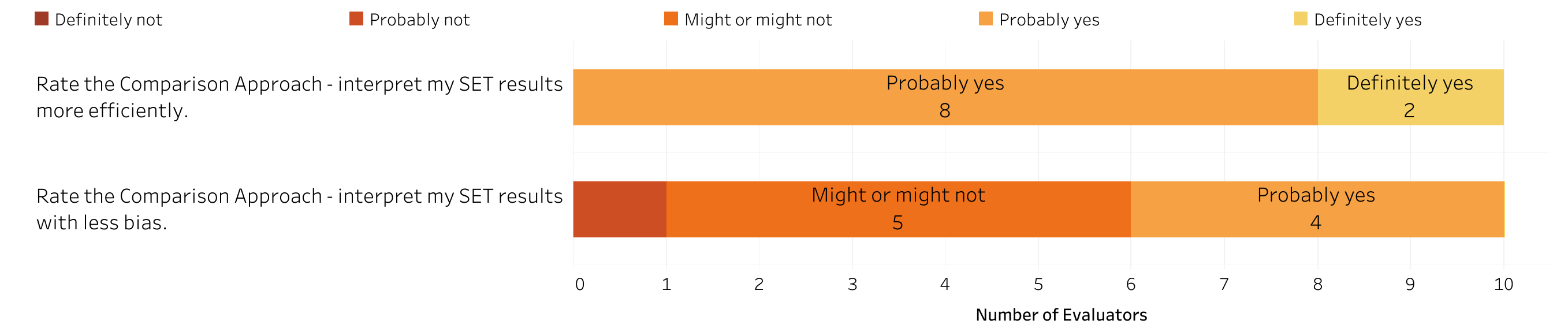}
         \subcaption{\textit{Overall, how useful is the \textbf{Comparison Approach} in summarizing students' opinions about your course and teaching?}}
         \label{fig:Q4-breakdown}
     \end{subfigure}
     \hfill
     \begin{subfigure}[b]{\textwidth}
         \centering
         \includegraphics[width=0.92\textwidth]{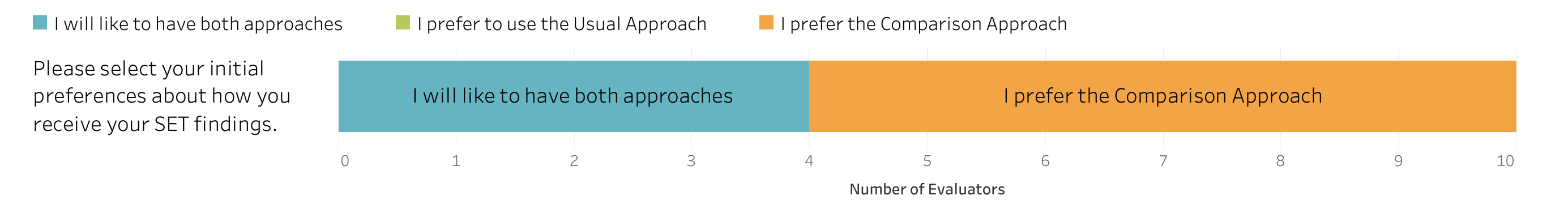}
         \subcaption{\textit{Overall, please select your initial preference about how you receive your SET findings.}}
         \label{fig:Q5-breakdown}
     \end{subfigure}
     \vspace{-15pt}
     \caption{Results of human evaluation comparing the standard SET report (the Usual Approach) and the \textsc{SETSum v1.0} website (the Comparison Approach).}
     \label{fig:HumanEval}
     \vspace{-15pt}
\end{figure*}

\begin{figure*}
    \centering
    \includegraphics[width=0.999\textwidth]{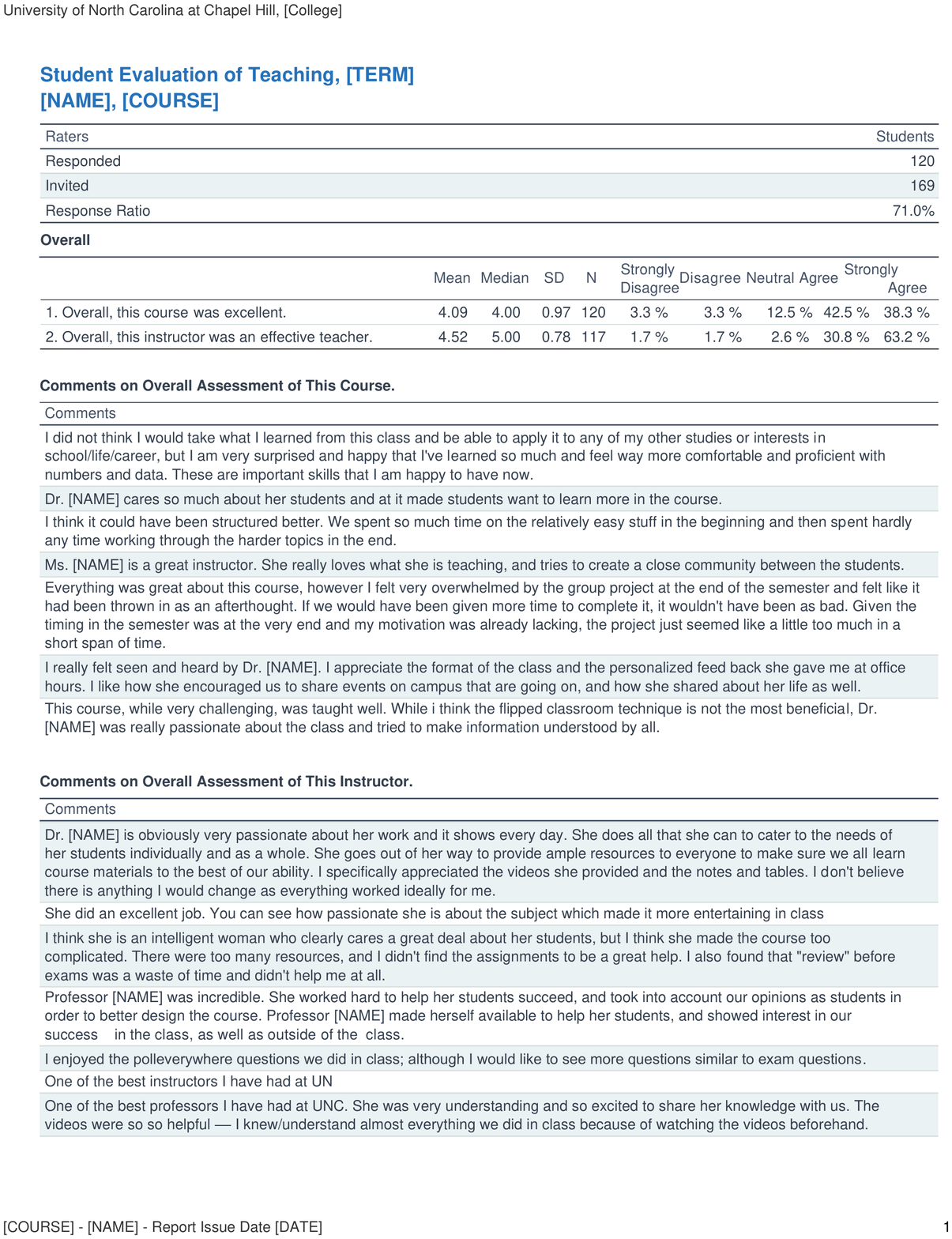}
    \vspace{-15pt}
    \caption{The standard PDF SET report.}
    \label{fig:usualSET}
    \vspace{-15pt}
\end{figure*}

\begin{figure*}
     \centering
     \begin{subfigure}[b]{\textwidth}
         \centering
         \includegraphics[width=0.99\textwidth]{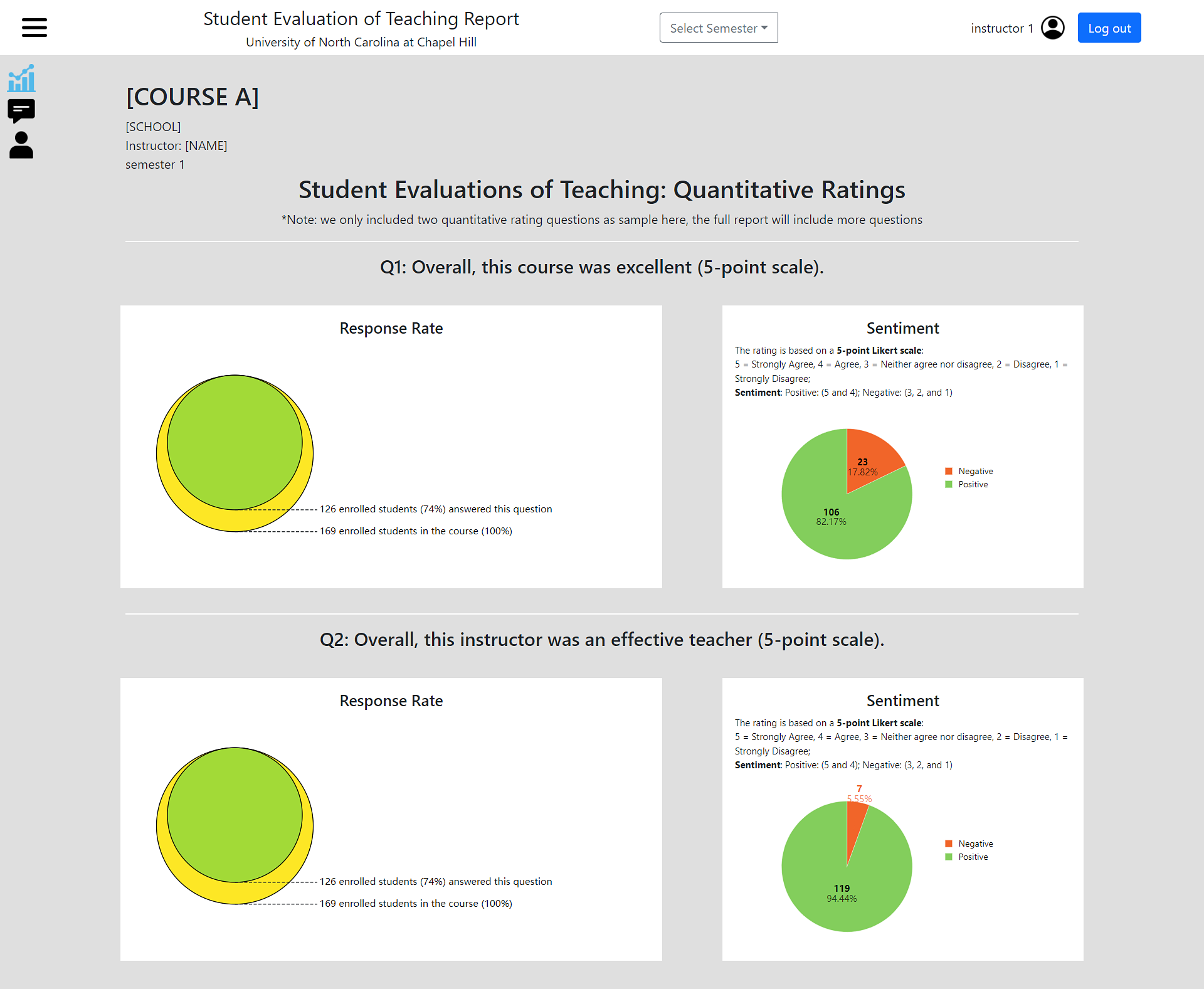}
         \subcaption{A page shows the Rating Analysis (Quantitative Questions) part.}
         \label{fig:SETSumWebv1-a}
     \end{subfigure}
     \vspace{-15pt}
     \caption{Screenshots of \textsc{SETSum} v1.0 (Part1, see Part2 in the next page).}
     \label{fig:SETSumWebv1}
     \vspace{-15pt}
\end{figure*}

\begin{figure*}
      \ContinuedFloat 
      \centering 
     \begin{subfigure}[b]{\textwidth}
         \centering
         \includegraphics[width=0.80\textwidth]{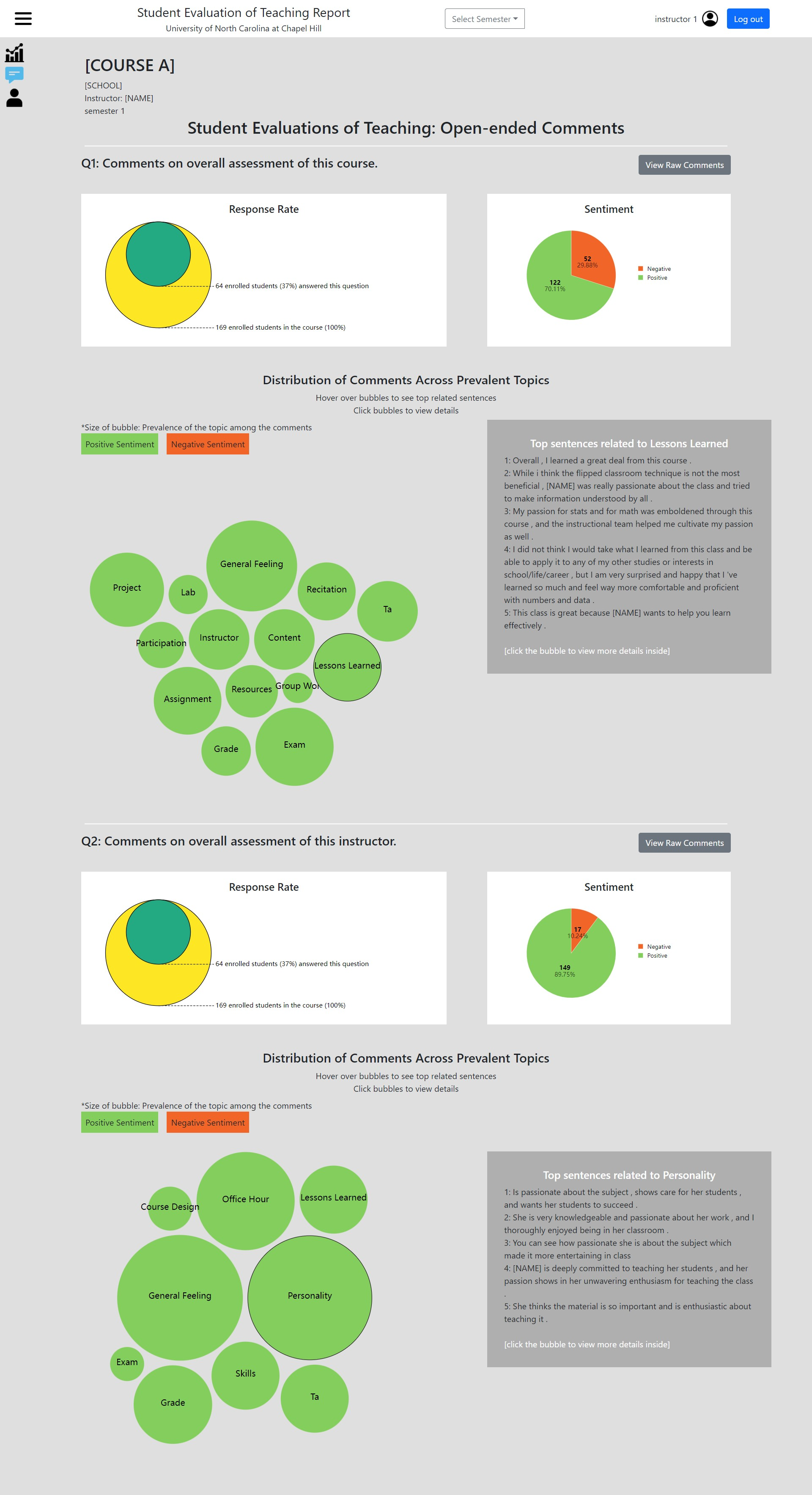}
         \subcaption{A page shows the Comments Analysis (Open-ended Questions) part.}
         \label{fig:SETSumWebv1-b}
     \end{subfigure}
     \vspace{-15pt}
     \caption{Screenshots of \textsc{SETSum} v1.0 (Part2).}
     \vspace{-15pt}
\end{figure*}

\begin{figure*}
     \centering
     \begin{subfigure}[b]{\textwidth}
         \centering
         \includegraphics[width=0.99\textwidth]{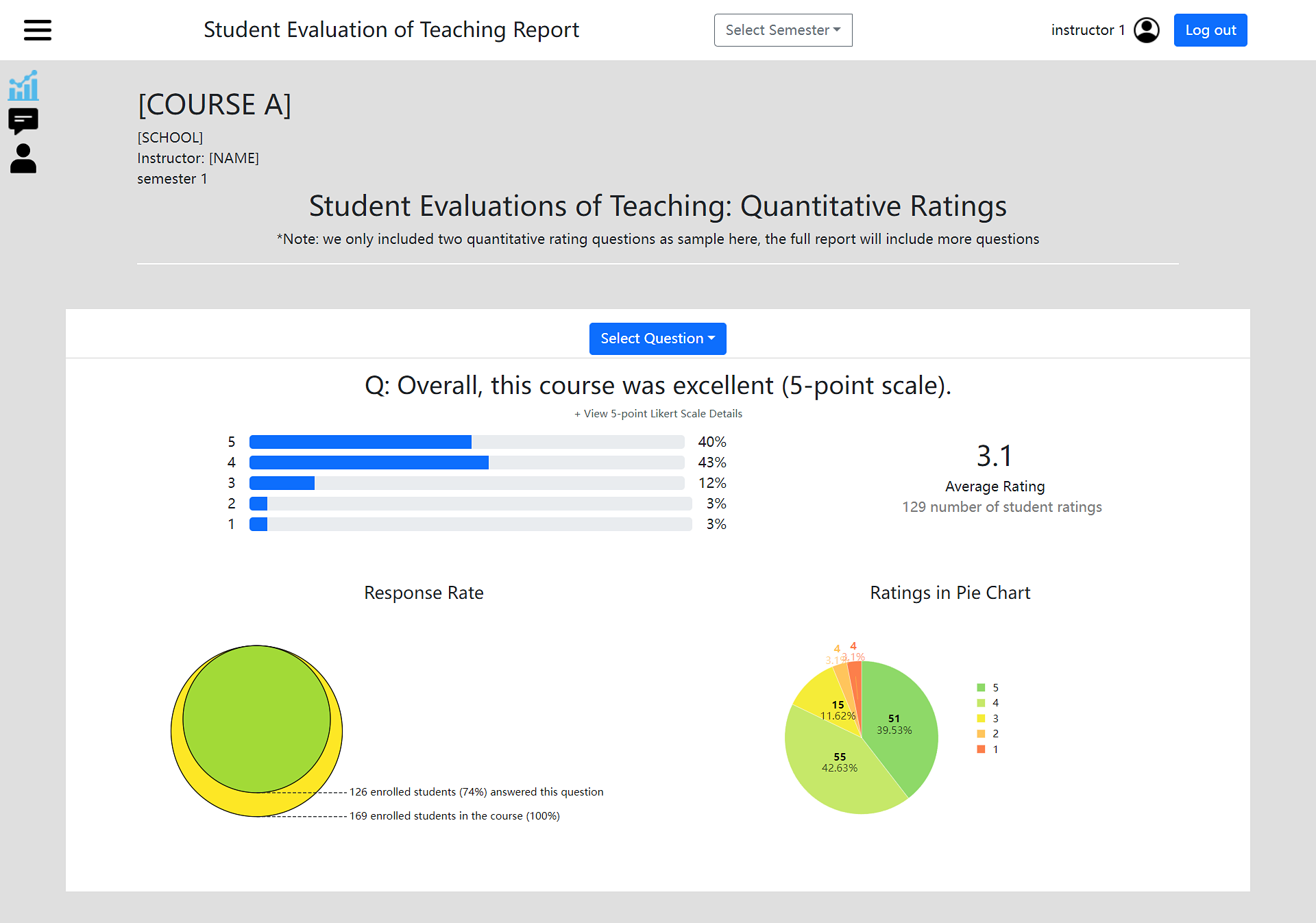}
         \subcaption{A page shows the Rating Analysis (Quantitative Questions) part.}
         \label{fig:SETSumWebv2-a}
     \end{subfigure}
     \vspace{-15pt}
     \caption{Screenshots of \textsc{SETSum} v1.1 (Part1, see Part2 in the next page).}
     \label{fig:SETSumWebv2}
     \vspace{-15pt}
\end{figure*}

\begin{figure*}
      \ContinuedFloat 
      \centering 
     \begin{subfigure}[b]{\textwidth}
         \centering
         \includegraphics[width=0.90\textwidth]{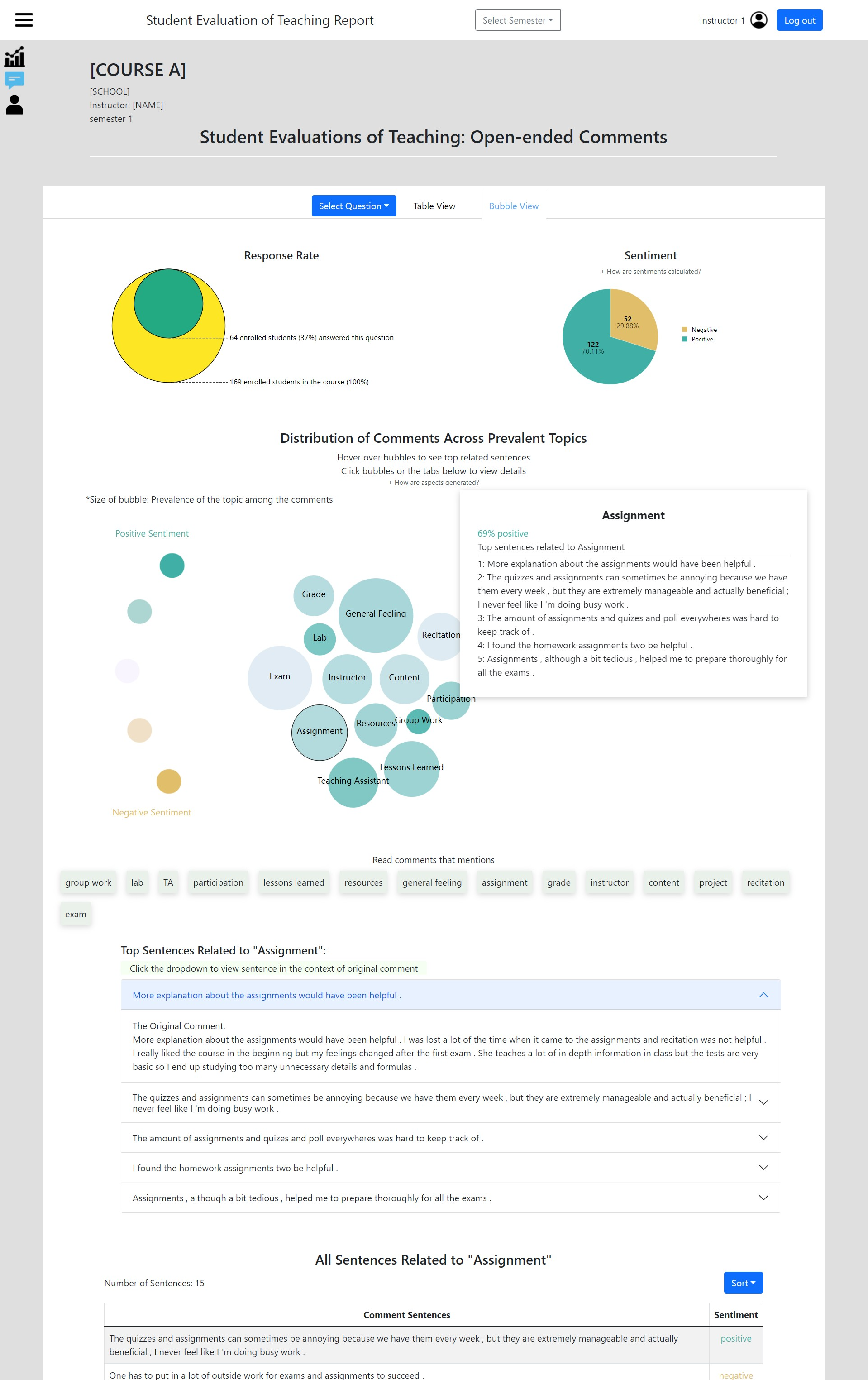}
         \subcaption{A page shows the Comments Analysis (Open-ended Questions) part.}
         \label{fig:SETSumWebv2-b}
     \end{subfigure}
     \vspace{-15pt}
     \caption{Screenshots of \textsc{SETSum} v1.1 (Part2).}
     \vspace{-15pt}
\end{figure*}

\begin{table*}[ht]
    \begin{subtable}[h]{0.99\textwidth}
        \centering
        \small
        \begin{tabular}{p{0.18\textwidth}|p{0.8\textwidth}}
        \toprule
        \textbf{Aspect} & \textbf{Top words (normalized weight)}\\
        \midrule
        assignment & assignment (0.33), homework (0.31), concept (0.17), reading (0.13), exercise (0.07) \\
        \midrule
        content & material (0.42), lecture (0.17), reading (0.15), subject (0.13), content (0.13) \\
        \midrule
        course design & syllabus (0.21), requirement (0.20), communicated (0.20), wish (0.20), discussion (0.19) \\
        \midrule
        exam & exam (0.31), test (0.24), question (0.20), answer (0.13), problem (0.13), \\
        \midrule
        general feeling & course (0.33), enjoyed (0.25), favorite (0.19), challenging (0.12), hard (0.11) \\
        \midrule
        grade & grading (0.39), feedback (0.19), harsh (0.16), midterm (0.16), easy (0.11) \\
        \midrule
        group work & group (0.30), project (0.24), recitation (0.20), work (0.06), team (0.20) \\
        \midrule
        instructor & professor (0.50), instructor (0.21), passionate (0.11), teach (0.11), condescending (0.06) \\
        \midrule
        lab & lab (0.32), hand (0.20), report(0.20), grading (0.10), experiment (0.18) \\
        \midrule
        lessons learned & learned (0.23), real (0.22), life (0.22), skill (0.19), understanding (0.14) \\
        \midrule
        participation & discussion(0.30), speak(0.23), comfortable (0.18), participation (0.16), stressful (0.13) \\
        \midrule
        project & project (0.30), instance (0.23),  expectation (0.18), clearly (0.16), explained (0.13) \\
        \midrule
        recitation & recitation (0.57), content (0.18), project (0.10), review (0.09), group (0.05) \\
        \midrule
        resources & peer (0.20), mentor (0.20), book (0.20), software (0.20), reference (0.20) \\
        \midrule
        teaching assistant (TA) & TA (0.42), job (0.20), helping (0.12), explained (0.06). available (0.20) \\
        \bottomrule
        \end{tabular}
       \caption{Highest ranked words list for each aspect of \textit{Comments on overall assessment of this course}.}
       \label{tab:week1}
    \end{subtable}
    \hfill
    \newline
    \vspace*{1em}
    \newline
    \begin{subtable}[h]{0.99\textwidth}
        \centering
        \small
        \begin{tabular}{p{0.18\textwidth}|p{0.8\textwidth}}
        \toprule
        \textbf{Aspect} & \textbf{Top words (normalized weight)}\\
        \midrule
        course design & lecture (0.28), assignment (0.21), topic (0.18), activity (0.17), structured (0.17) \\
        \midrule
        delivery & engaged (0.26), clear (0.22), lecture (0.22), example (0.16), explain (0.14)\\
        \midrule
        exam & unfair (0.25), fair (0.25), exam (0.23), guide (0.20), question (0.08) \\
        \midrule
        general feeling & professor (0.37), great (0.27), instructor (0.25), bad (0.05), overall (0.05) \\
        \midrule
        grade & grade (0.36), passing (0.20), average (0.20), exam (0.13), comment (0.11) \\
        \midrule
        lessons learned & conceptual (0.27), intellectual (0.27), learned (0.20), knowledge (0.16), understanding (0.11) \\
        \midrule
        office hour & office (0.38), hour (0.38), time (0.09), comment (0.08), meet (0.08) \\
        \midrule
        personality & enthusiastic (0.30), passionate (0.22), person (0.19), care (0.18), funny (0.12) \\
        \midrule
        recitation & recitation (0.26), time (0.14), project (0.20), group (0.20), organized (0.20) \\
        \midrule
        skills & knowledgeable (0.40), experience (0.26), information (0.14), quality (0.10), deep (0.10) \\
        \midrule
        teaching assistant (TA) & TA (0.41), interactive (0.15), supportive (0.15), constructive (0.15), feedback (0.15) \\
        \bottomrule
        \end{tabular}
        \caption{Highest ranked words list for each aspect of \textit{Comments on overall assessment of this instructor}.}
        \label{tab:week2}
     \end{subtable}
     \caption{Highest ranked words and normalized weight for each aspect.}
     \label{tab:aspects}
\end{table*}

\begin{table*}
\centering
\small
\resizebox{0.99\textwidth}{!}{%
\begin{tabular}{p{0.15\textwidth}|p{0.3\textwidth}|p{0.5\textwidth}}
\toprule
\textbf{Terminology} & \textbf{Description} & \textbf{Example}\\
\midrule
general feeling	& General high-level comments or overall feelings about the course & The course is really interesting for me as a CS major and I learned a lot form it. \\
\midrule
instructor & Any comments towards the instructor & Professor [NAME] is a joy, and is incredibly understanding, passionate, and enjoyable to simply listen to in class! \\
\midrule
teaching assistant (TA) & Any comments related to TA	& Resources are always available, the instructors and TAs were easily accessible and always friendly, the material was challenging, and examples were always fun and engaging. \\
\midrule
lab	& Any comments related to lab & I thought this course was very engaging and I liked that it was very hands on, like a lab should be. \\
\midrule
recitation & Any comments related to recitation	& This recitation was a bit odd. \\
\midrule
course design & Any comments on the organization and structure of the course & I thought this course was excellently structured and formatted. \\
\midrule
assignment	& Any comments related to homework/assignments & The assignment is really, really well designed that it builds upon each other from assignment 2 through assignment 9 and it helped me exercise various topics/concepts that I learned from class.\\
\midrule
exam & Any comments related to exam/test & This class is extremely hard and the second test is expected to be failed by most students, which is ridiculous. \\
\midrule
content	& Any comments related to course materials or specific contents of the course & The material was very useful for our course  although the professors could have made a better connection with the techniques learned in the lab. \\
\midrule
participation	& Talk about the participation / attendance / engagement / discussion	& Needs more class participation and discussion. \\
\midrule 
grade & Comments on the grading of the course &	Harsh grading on lab reports. \\
\midrule
group work	& Any comments related to group work & All of the recitations consisted of group work towards a final project, though the early recitations seemed largely irrelevant to the project. \\
\midrule
resources & Resources provided by course such as readings, textbooks, peer tutors etc. & I did however get all the help I needed from the peer mentors. \\
\midrule
lessons learned	& Learning outcomes or skills acquired from the course	& The professor is really good, I learned a lot of interesting and classic dramas this semester. \\
\midrule
project	& Any comments related to projects & The projects were interesting, but I can think of at least one instance where the expectations of the project were not explained clearly enough.  \\
\bottomrule
\end{tabular}
}
\caption{Aspect Annotation Terminology for \textit{Comments on overall assessment of this course}.}
\label{tab:terminology_cou}
\end{table*}

\begin{table*}
\centering
\small
\resizebox{0.99\textwidth}{!}{%
\begin{tabular}{p{0.15\textwidth}|p{0.3\textwidth}|p{0.5\textwidth}}
\toprule
\textbf{Terminology} & \textbf{Description} & \textbf{Example}\\
\midrule
general feeling	& General high-level comments or overall feelings about the instructor & Awesome Professor. \\
\midrule
teaching assistant (TA) & Any comments related to TA & One of the best TAs I have had so far at UNC \\
\midrule
recitation & Any comments related to recitation & Recitation felt like a waste of time. \\
\midrule
office hour	& Any comments related to office hour & She was great during her office hours and was always concerned that we understood the material. \\
\midrule
personality	& Describe personality of the instructor & She cares a lot about the subject material and her students. \\
\midrule
skills & Describe the skill sets or experiences of the instructor & The instructor had a deep understanding of the course material and provided many real world examples built from her own experience and previous work. \\
\midrule
grade & Comments on the grading style & The instructor was clear at explaining information and fairly evaluating all assignments. \\
\midrule
delivery & How the instructor delivers the information and explains concepts & I really enjoyed her teaching style, she helped us through tough topics by breaking them down into more digestible chunks and was really positive overall, which helped for class moral. \\
\midrule
course design & Comments on the organization and structure of the course & I didn't really get to know the TA because we didn't have a lot of recitations. \\
\midrule
lessons learned	& Learning outcomes or skills acquired from the course & I now have a greater understanding of the German language and of Swiss;German literature and culture.  \\
\midrule
exam & Any comments related to exam/test & Exams were hard, as expected, but I think some questions were unfair at times. \\
\bottomrule
\end{tabular}
}
\caption{Aspect Annotation Terminology for \textit{Comments on overall assessment of this instructor}.}
\label{tab:terminology_ins}
\end{table*}

\end{document}